\documentclass[conference,compsoc]{IEEEtran}

\usepackage{cite}
\usepackage{amsmath,amssymb}
\usepackage{graphicx}
\usepackage{textcomp}
\usepackage{xcolor}
\usepackage{booktabs}
\usepackage{url}
\usepackage{microtype}
\usepackage{gensymb}

\graphicspath{{figures/}}
\newcommand{\degc}{\,\textdegree C}

\begin{document}

\title{Thermal-Headroom Routing for Efficient Robot Perception in Extreme Heat}

\author{
\IEEEauthorblockN{Sparsh Roy\IEEEauthorrefmark{1}\IEEEauthorrefmark{2}\IEEEauthorrefmark{3},
Kevin Zhou\IEEEauthorrefmark{2},
Arhaan Shah\IEEEauthorrefmark{4},
Shreyan Garg\IEEEauthorrefmark{4},\\
Vihaan Joshi-Apte\IEEEauthorrefmark{4},
Maxim Kudryashov\IEEEauthorrefmark{2},
Nanda Cheruvanthala\IEEEauthorrefmark{4}}
\IEEEauthorblockA{\IEEEauthorrefmark{1}Massachusetts Institute of Technology, Cambridge, MA, USA}
\IEEEauthorblockA{\IEEEauthorrefmark{2}Hopewell Valley Central High School, Pennington, NJ, USA}
\IEEEauthorblockA{\IEEEauthorrefmark{4}Paradise Valley High School, Phoenix, AZ, USA}
\IEEEauthorblockA{\IEEEauthorrefmark{3}Corresponding author}
\IEEEauthorblockA{\small sparshr@mit.edu (S.~Roy), k.zhou@mit.edu (K.~Zhou), arhaan@mit.edu (A.~Shah),\\
shreyangarg@gmail.com (S.~Garg), vjoshiapte@gmail.com (V.~Joshi-Apte),\\
maximkud21+research@gmail.com (M.~Kudryashov), nanda.vadakara@gmail.com (N.~Cheruvanthala)}
}

\maketitle

\begin{abstract}
A robot working in a Phoenix summer runs its detector on a small
embedded processor that heats itself. Once the chip reaches its thermal
limit, the governor throttles it and every detector configuration slows
down together. Adaptive inference routers pick a model size and input
resolution per frame and react to latency, which in hot weather arrives
after the router's own earlier choices have already heated the chip. We
model this with an RC thermal plant wired in closed loop to a
deadline-constrained router over four YOLO11 configurations, and
propose Thermal-Headroom Routing (THR), which forecasts the chip
temperature in closed form and skips configurations that would cross
the throttle threshold. Heating measured on an Apple M5 under nonstop
YOLO11 inference fits the one-node model within 1\degc, with a time
constant and thermal resistance close to the values we assumed. We test
THR on seven real traces with camera
degradation, eight ambient temperatures and 20 seeds, with Holm
correction across traces, against a latency-belief router, a
fixed-threshold thermostat, duty caps tuned offline, and a control
given THR's own heavy-model budget. With the deadline at the large
model's median latency and the air between 45 and 55\degc, THR never
throttles and beats the belief router by 3.9 to 7.5 percentage points
of deadline misses and 12 to 15\% of energy per on-time frame on every
trace. All temperature-aware policies, THR included, lie on one
accuracy-timeliness curve set by the share of heavy-model frames, so at
one ambient on a known chip a well-tuned thermostat matches THR. The
forecast pays off when conditions move away from the ones a rule was
tuned for. A duty cap tuned at 45\degc{} throttles 16\% of the time at
50\degc{} and 31\% at 55\degc, and on a chip 20\% hotter than assumed an
88\degc{} thermostat throttles up to 29\% of the time, while THR,
given that chip's measured heating parameters, stays at zero. These
results hold, with larger effects, when every decision is given one
deadline period of wall-clock time. With 25\% deadline slack,
throttling costs little and the belief router wins on on-time utility. Code and
outputs are public.%
\footnote{\url{https://github.com/ImSpxrsh/thermal-headroom-routing}}
\end{abstract}


\section{Introduction}

\begin{figure}[t]
\centering
\includegraphics[width=\columnwidth]{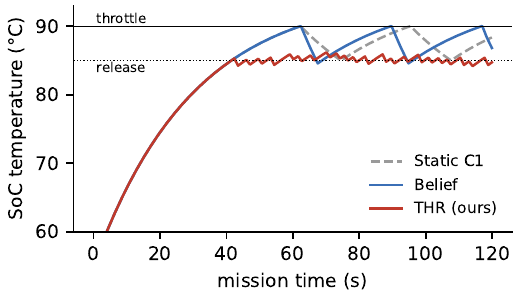}
\caption{SoC temperature during one 2-minute mission at 45\degc{} (fog
trace, seed 0). Static C1 and Belief go through repeated throttle
cycles. THR alternates C1 and C2 and stays just below the release
threshold.}
\label{fig:trace}
\end{figure}

Consider a Jetson-class module in a sealed enclosure on a robot in the
Sonoran Desert, with the air at 45\degc. If the perception stack runs
its largest detector on every frame, the chip reaches its throttle
limit in about a minute. The governor then down-clocks it, and the
small models slow down along with the large one until the chip cools
\cite{bhat2019power,benoitcattin2020throttling}. In our simulations a
policy that always runs the large model goes 12.5\,s without an on-time
detection at 45\degc{} and 91\,s at 55\degc. Fig.~\ref{fig:trace} shows one
two-minute mission at 45\degc: a router that reacts to latency drives
the chip into repeated throttle cycles, and THR keeps it just under the
line.

Adaptive inference is the usual answer to a tight latency budget.
Early-exit networks and routers choose cheaper computation per input or
per frame \cite{teerapittayanon2016branchynet,laskaridis2021adaptive,wang2021notallimages},
and contention-aware detectors such as ApproxDet \cite{xu2020approxdet}
also watch the load from co-located applications. These systems react
to latency once it has risen. On a hot chip that signal arrives late,
because the slowdown was caused by earlier choices and lasts tens of
seconds. Thermal-aware DVFS \cite{kim2021ztt} and thermal-aware
multi-DNN scheduling on phones \cite{jeon2026phoenix} address heat by
controlling frequency or placement. We ask a narrower question that
matters for robots with a fixed set of detector configurations: how
should the router itself use temperature, and how much does a
temperature forecast add over simpler rules? Belief-space perception
routing \cite{roy2026beliefspace}, which this work extends, left
thermal effects out of its contention model.

Our contributions:
(i) a closed-loop model of thermal contention for adaptive inference,
in which the router's configuration choice sets the power input to an
RC thermal node with a hysteretic throttle (\S\ref{sec:model});
(ii) THR, a closed-form rule that keeps the forecast temperature under
the throttle limit at about 7\,$\mu$s per frame (\S\ref{sec:thr});
(iii) an evaluation against baselines chosen to beat it: a thermostat,
duty caps tuned offline, a matched-budget control in the spirit of the
matched-input controls of \cite{zhou2026matchedinput}, a deadline
sweep, and five variations of the true chip, with 20 seeds and Holm
correction (\S\ref{sec:results}); and
(iv) an account of where THR helps and where it does not. Its gain
over latency-only routing comes from rationing the heavy model. The
forecast itself pays off when the chip heats differently from what a
fixed rule assumed, and with deadline slack, throttling costs less than
rationing does.

\section{Related Work}

\textit{Adaptive and efficient inference.} Early-exit networks
\cite{teerapittayanon2016branchynet,laskaridis2021adaptive}, dynamic
transformers \cite{wang2021notallimages} and early exiting for driving
\cite{adee2025,navee2025,temporalpatience2023} vary computation with
the input. Modality gating \cite{dynmm,mees2016choosing,pramr2025}
decides which sensors to process. ApproxDet \cite{xu2020approxdet}
selects detection branches under a latency SLA using measured
contention from co-located apps. In all of these the compute budget is
fixed or observed. None treats it as a state the policy's own choices
drive.

\textit{Thermal behaviour of edge inference.} On a Raspberry Pi, long
CNN runs throttle within seconds without active cooling, and outdoor
ambients between 22 and 36\degc{} change throughput by up to 27.7\%
\cite{benoitcattin2020throttling}. A Jetson AGX Orin in an outdoor
enclosure throttled under a synthetic stress test at a +40\degc{}
setpoint (GPU 95.6\degc, CPU 99.0\degc), while a YOLOv8 workload drew
about half the power and stayed below the threshold
\cite{krislauks2026orin}. Throttling therefore depends on the model,
the cooling and the duty cycle, which is why we sweep the plant
parameters below. Measurements on phone SoCs show throttling dominating
sustained performance \cite{bhat2019power}.

\textit{Thermal management.} HotSpot-style RC models
\cite{skadron2003hotspot} and temperature-aware MPSoC scheduling
\cite{coskun2007temperature} predict hot spots and schedule around
them. zTT \cite{kim2021ztt} learns CPU/GPU frequency settings with
reinforcement learning so that phones avoid throttling. Phoenix
\cite{jeon2026phoenix} combines processor placement with multi-exit
networks to delay throttling for multi-DNN video apps. THR works one
level up: it leaves the governor alone and chooses among detector
configurations using a temperature forecast. It could run on top of a
DVFS policy like zTT.

\textit{Sensor- and compute-aware routing.} Belief-space perception
routing \cite{roy2026beliefspace} tracks blur, illumination and
occlusion faults per camera with switching HMMs, keeps a latency-based
contention belief, and fuses the two with a noisy-OR term. We reuse its
fault beliefs, frontier and traces from TartanDrive
\cite{triest2022tartandrive,sivaprakasam2024tartandrive2}, RADIATE
\cite{sheeny2021radiate} and WoodScape \cite{yogamani2019woodscape}.

\section{System Model}
\label{sec:model}

\textit{Frontier and deadline.} There are four configurations, from most
to least expensive: C1 is YOLO11x at 1280\,px, C2 is YOLO11x at 640, C3
is YOLO11n at 1280 and C4 is YOLO11n at 640, with per-frame latencies
profiled under normal and contended compute \cite{roy2026beliefspace}.
The base deadline $D$ is C1's median latency under normal compute,
between 211 and 341\,ms depending on the trace, so C1 meets $D$ about
half the time on a cool chip and C2 to C4 always meet it. Section
\ref{sec:slack} relaxes $D$ to $1.25\times$ and $1.5\times$. Accuracy is
the IoU-matched F1 agreement between a configuration's detections and
C1's on the same frame. This pseudo-ground truth ranks configurations
differently from real annotations \cite{roy2026beliefspace}, so
accuracy and utility numbers are internal consistency checks. Miss,
throttling and energy numbers do not depend on it.

\textit{Thermal plant.} The SoC is one thermal node with time constant
$\tau$ and junction-to-ambient resistance $R_{\mathrm{th}}$:
\begin{equation}
T_{t+1} = T_t + \tfrac{\Delta t}{\tau}\big(T_{\mathrm{amb}} + R_{\mathrm{th}}\,P_t - T_t\big).
\label{eq:rc}
\end{equation}
Power depends on the configuration $c_t$ served at frame $t$:
$P_t = s_t\,[P_{\mathrm{idle}} + (P_{\max}-P_{\mathrm{idle}})\,u(c_t)]$,
where $u(c)=\min(1,\tilde{\ell}_c/D_0)$ is the configuration's median
normal-compute latency as a fraction of the base deadline $D_0$ (so a
deadline sweep does not change power), and $s_t$ is 0.6 while
throttled and 1 otherwise. The governor starts throttling at
$T\ge T_{\mathrm{hi}}$ and stops at $T\le T_{\mathrm{lo}}$. While it
throttles, every configuration's latency is drawn from its profiled
contended distribution, whose p95 is about $2.5\times$ the normal one.
Table~\ref{tab:params} lists the parameters. They describe a sealed,
sun-loaded enclosure running the large model continuously: running C1
nonstop settles at $T_{\mathrm{amb}}+50$\degc, so throttling starts once
the air is above 40\degc. This is harsher than the well-ventilated Orin
with a lighter YOLOv8 model in \cite{krislauks2026orin}, and we chose
the values ourselves. Section~\ref{sec:plant} varies the true plant,
and Section~\ref{sec:m5} compares it with heating measured on an
Apple M5.

\begin{table}[t]
\caption{Thermal-plant parameters (fixed before any run).}
\label{tab:params}
\centering\footnotesize
\begin{tabular}{@{}lc@{\hspace{10pt}}lc@{}}
\toprule
$\Delta t$ (frame) & 0.1\,s & $P_{\mathrm{idle}}$ / $P_{\max}$ & 5 / 25\,W \\
$\tau$ & 30\,s & $T_{\mathrm{hi}}$ / $T_{\mathrm{lo}}$ & 90 / 85\degc \\
$R_{\mathrm{th}}$ & 2.0\,\textdegree C/W & throttled power & $0.6\times$ \\
\bottomrule
\end{tabular}
\end{table}

\textit{Closed loop.} Whether the chip is throttled at frame $t$
depends on what the router chose before $t$, so each policy runs in
closed loop with the plant. For a given seed, all policies see the same
pre-drawn latency samples (normal and contended), the same camera
measurement noise and the same temperature-sensor noise.

\textit{Frame timing.} In the main experiments the traces play at their
10\,Hz recording rate and each recorded frame is one routing decision.
The chosen configuration runs once on that frame, its latency is drawn
from the profiled distribution, and the result counts as a miss if it
exceeds $D$. Late results are dropped, so there is no queue and a late
frame never delays the next one. Power is charged through the duty
$u(c)$, and the thermal model advances $\Delta t=0.1$\,s per decision.
Because C1 takes about $D$ (211 to 341\,ms) to run, this compresses
wall-clock time by a factor of 2.1 to 3.4. Section~\ref{sec:timing}
repeats the heat experiment with physically consistent timing, where
every decision lasts exactly $D$, and the conclusions hold.

\section{Policies}
\label{sec:thr}

All policies use the objective from \cite{roy2026beliefspace}. They
predict a compute state, keep the configurations whose median latency
in that state meets $D$, and serve the one with the highest expected
accuracy given the camera-fault belief $b_f$. A 10-frame dwell keeps
them from switching back and forth.

\textit{Online compute belief.} In closed loop the router learns only
the latency of the configuration it served, one frame later. A
two-state forward filter (stay probabilities 0.95 and 0.90) runs on one
binary observation, whether that latency exceeded $1.5\times$ the
configuration's normal median, and outputs $b_c$.

\textit{THR.} With the on-die reading $\hat T$ ($\sigma=1$\degc{} noise),
holding configuration $c$ for $H$ frames gives, from
Eq.~\eqref{eq:rc},
\begin{equation}
\hat T_{H}(c) = T_{\infty}(c) + \big(\hat T - T_{\infty}(c)\big)\big(1-\tfrac{\Delta t}{\tau}\big)^{H},
\label{eq:forecast}
\end{equation}
with $T_{\infty}(c)=T_{\mathrm{amb}}+R_{\mathrm{th}}P(c)$. THR drops $c$
when $\hat T_H(c) \ge T_{\mathrm{hi}}-m$, with $H=50$ frames (5\,s) and
$m=2$\degc. C4 is always allowed. Each frame costs four evaluations of
Eq.~\eqref{eq:forecast}; with the belief update and selection we
measured 7.3\,$\mu$s per frame in Python on one Apple M5 core, under
0.004\% of one C1 inference.

\textit{Baselines.} Static C1 and Static C4 always serve one
configuration. Reactive uses latency only. Belief is the decoupled
router of \cite{roy2026beliefspace} with the online $b_c$. Thermostat
is Belief with C1 barred whenever $\hat T\ge88$\degc, the same line THR
forecasts against, so the two differ only in forecasting. Cap (45\degc)
and Cap (55\degc) are Belief with a running cap on the share of C1
frames, set to the share THR used at that ambient. They stand for a
duty cap tuned once in the lab. Matched budget gets THR's own C1 share
for each seed and ambient, which tests whether THR helps by timing its
C1 frames or only by using fewer of them. For heat combined with camera
faults, Joint adds the noisy-OR fault-coupling term of
\cite{roy2026beliefspace}, $p_{\mathrm{cont}} = 1-(1-b_c)(1-\kappa b_f)$,
and Joint\,+\,THR combines it with THR.

\section{Experimental Setup}

\textit{Traces.} Seven real traces with the frontiers and fault labels
of \cite{roy2026beliefspace}: the full TartanDrive drive, which has
no camera faults and is our negative control, a 309-frame
stretch of rough terrain, RADIATE rain, snow, fog and night, and
WoodScape fisheye video with real soiling masks. Each trace is repeated
to fill a 1200-decision mission (2 minutes at the 10\,Hz recording
rate).

\textit{Experiments.} E1 re-scores the original fault-coupled contention
schedule for bursts of misses. E2 sweeps ambient temperature from 20 to
55\degc{} with thermal contention only. E3 adds fault-coupled
contention on top of throttling. E4 gives THR a wrong $R_{\mathrm{th}}$
or $\tau$. E5 relaxes the deadline. E6 changes the true chip. E7 and E8
vary THR's $H$ and $m$ and the thermostat threshold, as sensitivity
analyses. E9 sweeps fixed C1 caps between 0.2 and 1.0. E11 repeats
E2 with physically consistent frame timing (\S\ref{sec:timing}).

\textit{Protocol.} Following the validation-locked protocol of
\cite{zhou2026matchedinput}, every setting behind the headline numbers
(plant parameters, $H$, $m$, dwell, the $1.5\times$ slow threshold, the
88\degc{} thermostat) was fixed in code before the first evaluation
run, and each result file records
\texttt{eval\_labels\_used\_for\_selection: false}. Every simulated
table and figure is regenerated from the released code by
\texttt{run\_extreme\_env.py} followed by
\texttt{make\_extreme\_figures.py}, which takes about 8 minutes on one
laptop core. The caps tuned at
45 and 55\degc{} are fit on the evaluation seeds, which favours them.
We use 20 seeds, which change latency draws, camera noise and
temperature-sensor noise on the same real frames. Each comparison is a
paired $t$-test on per-seed differences, and $^\ast$ marks a difference
that survives Holm correction across the seven traces of its family at
$\alpha=0.05$.

\section{Results}
\label{sec:results}

\begin{figure*}[t]
\centering
\includegraphics[width=\textwidth]{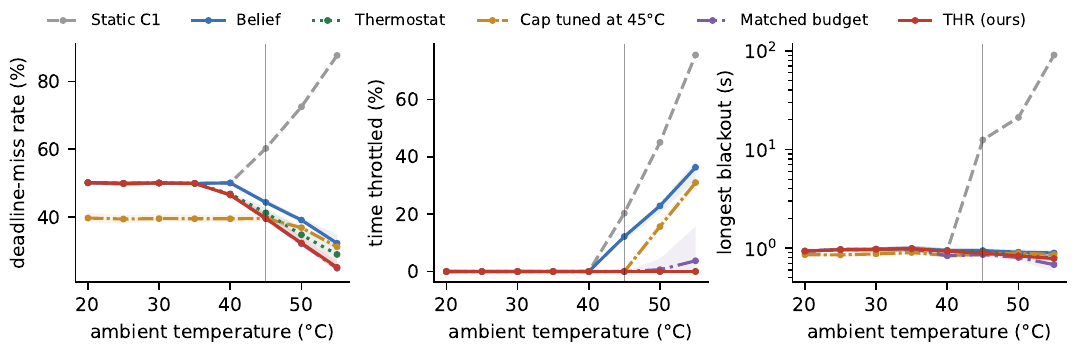}
\caption{Ambient sweep (E2), base deadline. Lines are the mean over
seven traces; bands span the lowest and highest trace. Left:
deadline-miss rate. Middle: share of the mission throttled. Right:
longest run of consecutive misses (log scale). THR (red) and the
thermostat never throttle. The cap tuned at 45\degc{} wastes accuracy
below 45\degc{} (lower miss, Table~\ref{tab:caps}) and throttles above
it.}
\label{fig:sweep}
\end{figure*}

\subsection{Cool air}
At 35\degc{} and below no policy brings the chip to $T_{\mathrm{hi}}$.
THR, Belief, Reactive, Thermostat and Matched budget make the same
choice on every frame and their miss rates differ by exactly 0.0
percentage points on all seven traces (Fig.~\ref{fig:sweep}). At
40\degc{} THR starts holding C1 back a little early (C1 on 93 to 94\% of
frames, agreement 0.97 to 1.00), because its 5\,s forecast already
crosses 88\degc.

\subsection{Hot air, against latency-only routing}
Table~\ref{tab:eff} lists miss rate, throttling, blackout length,
power, energy and accuracy for every policy. Static C1 is throttled
20\% of the time at 45\degc{} with a 12.5\,s blackout, and 76\% of the
time at 55\degc{} with a 91\,s blackout and 88\% of deadlines missed.
Belief and Reactive keep blackouts near 1\,s but are throttled 11 to
13\% of the mission at 45\degc{} and 32 to 38\% at 55\degc, because they
react after the governor has stepped in (Fig.~\ref{fig:trace}).

THR was never throttled on any trace, seed or temperature. Against
Belief it lowers the miss rate by 3.9 to 5.4 points at 45\degc{} and
6.7 to 7.5 at 55\degc, significant on every trace at both temperatures
(Table~\ref{tab:heat}). Energy per on-time frame falls by 12\% at
45\degc{} and 13\% at 55\degc, also on every trace. The price is
accuracy: mean agreement at 45\degc{} drops to 0.950, against 0.975
for Belief. On-time
utility (accuracy if on time, else zero) still rises significantly on
5 of 7 traces at 45\degc{} and 6 of 7 at 55\degc. Static C4 beats every
policy on timing and energy at an agreement of 0.695.

\begin{table}[t]
\caption{Efficiency, quality and cost in E2 (base deadline), mean over
7 traces and 20 seeds. Miss and Thr.\ (throttled) in \%. Blackout is
the longest run of consecutive misses (s). $P$ is mean SoC power (W)
and $E_{\mathrm{ot}}$ is energy per on-time frame (J). Acc.\ is pseudo-GT
agreement.}
\label{tab:eff}
\centering\footnotesize
\setlength{\tabcolsep}{3.2pt}
\begin{tabular}{@{}lcccccc@{}}
\toprule
Policy & Miss & Thr. & Blackout & $P$ & $E_{\mathrm{ot}}$ & Acc. \\
\midrule
\multicolumn{7}{l}{\textit{Ambient 45\degc}} \\
Static C1 & 60.3 & 20.3 & 12.5 & 23.0 & 5.79 & 1.000 \\
Static C4 & 0.0 & 0.0 & 0.0 & 5.6 & 0.56 & 0.695 \\
Reactive & 44.2 & 12.4 & 0.9 & 22.6 & 4.06 & 0.975 \\
Belief & 44.3 & 12.2 & 0.9 & 22.6 & 4.07 & 0.975 \\
Thermostat & 41.2 & 0.0 & 0.9 & 22.2 & 3.79 & 0.958 \\
Cap (45\degc) & 39.5 & 0.0 & 0.9 & 21.9 & 3.62 & 0.946 \\
Cap (55\degc) & 25.1 & 0.0 & 0.7 & 17.6 & 2.36 & 0.879 \\
Matched budget & 39.6 & 0.0 & 0.9 & 21.9 & 3.63 & 0.948 \\
THR (ours) & 39.6 & 0.0 & 0.9 & 21.7 & 3.60 & 0.950 \\
\multicolumn{7}{l}{\textit{Ambient 55\degc}} \\
Static C1 & 87.7 & 75.5 & 90.7 & 17.4 & 14.22 & 1.000 \\
Static C4 & 0.0 & 0.0 & 0.0 & 5.6 & 0.56 & 0.695 \\
Reactive & 32.0 & 37.3 & 0.9 & 17.9 & 2.63 & 0.927 \\
Belief & 32.3 & 36.3 & 0.9 & 17.9 & 2.64 & 0.929 \\
Thermostat & 29.0 & 0.0 & 0.8 & 18.4 & 2.59 & 0.899 \\
Cap (45\degc) & 31.2 & 31.0 & 0.9 & 17.9 & 2.60 & 0.902 \\
Cap (55\degc) & 24.9 & 3.9 & 0.7 & 17.4 & 2.31 & 0.877 \\
Matched budget & 24.9 & 3.7 & 0.7 & 17.4 & 2.31 & 0.878 \\
THR (ours) & 25.1 & 0.0 & 0.8 & 17.2 & 2.30 & 0.883 \\
\bottomrule
\end{tabular}
\end{table}

\begin{table*}[t]
\caption{Heat regime (E2) per trace, base deadline. Blackout C1 is Static
C1's longest run of consecutive misses (s). Thr.\ is the \% of the
mission throttled. $\Delta$ THR is the reduction in miss rate relative
to Belief (pp). THR$-$Tst and THR$-$MB are THR's miss-rate advantage over
the 88\degc{} thermostat and the matched-budget control (positive means
THR misses less). $^\ast$: Holm-significant across the seven traces.}
\label{tab:heat}
\centering\footnotesize
\setlength{\tabcolsep}{3pt}
\resizebox{\textwidth}{!}{%
\begin{tabular}{@{}l cccccc c cccccc@{}}
\toprule
& \multicolumn{6}{c}{Ambient 45\degc} && \multicolumn{6}{c}{Ambient 55\degc} \\
\cmidrule{2-7}\cmidrule{9-14}
Trace & Blackout C1 & Thr.\ Belief & Thr.\ THR & $\Delta$ THR & THR$-$Tst & THR$-$MB &&
Blackout C1 & Thr.\ Belief & Thr.\ THR & $\Delta$ THR & THR$-$Tst & THR$-$MB \\
\midrule
Off-road (clean) & 12.5 & 13 & 0 & +4.9$^\ast$ & +1.6$^\ast$ & +0.2 &  & 90.7 & 38 & 0 & +7.2$^\ast$ & +3.9$^\ast$ & -0.1 \\
Off-road seg. & 12.5 & 12 & 0 & +4.8$^\ast$ & +1.6$^\ast$ & -0.2 &  & 90.7 & 37 & 0 & +6.8$^\ast$ & +3.8$^\ast$ & -0.3 \\
Rain & 12.5 & 13 & 0 & +4.4$^\ast$ & +1.9$^\ast$ & -0.1 &  & 90.8 & 36 & 0 & +7.1$^\ast$ & +3.7$^\ast$ & -0.3 \\
Snow & 12.5 & 12 & 0 & +5.4$^\ast$ & +1.7$^\ast$ & +0.1 &  & 90.7 & 38 & 0 & +7.2$^\ast$ & +4.1$^\ast$ & -0.1 \\
Fog & 12.4 & 11 & 0 & +4.8$^\ast$ & +1.9$^\ast$ & +0.0 &  & 90.7 & 34 & 0 & +7.5$^\ast$ & +3.8$^\ast$ & +0.0 \\
Night & 12.6 & 13 & 0 & +5.0$^\ast$ & +1.5$^\ast$ & +0.1 &  & 90.7 & 38 & 0 & +7.3$^\ast$ & +4.4$^\ast$ & -0.0 \\
Soiling (fisheye) & 12.5 & 11 & 0 & +3.9$^\ast$ & +1.3$^\ast$ & -0.1 &  & 90.7 & 32 & 0 & +6.7$^\ast$ & +2.9$^\ast$ & -1.1$^\ast$ \\
\bottomrule
\end{tabular}}
\end{table*}

\subsection{Simple temperature rules and the trade-off curve}
\label{sec:simple}
\begin{figure*}[t]
\centering
\includegraphics[width=0.86\textwidth]{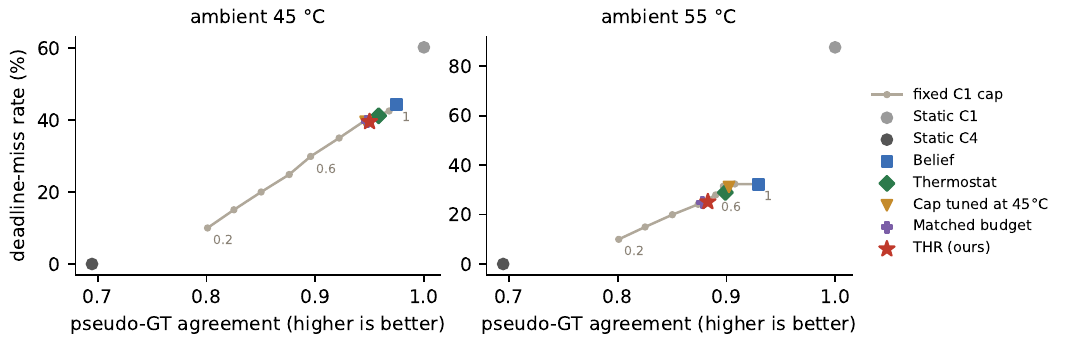}
\caption{Accuracy against deadline-miss rate (E2 and E9), mean over
seven traces. The grey line is a fixed C1 cap between 0.2 and 1.0. Every
temperature-aware policy lies on it; Belief sits off the curve at
55\degc{} because throttling raises its miss rate.}
\label{fig:pareto}
\end{figure*}

A thermostat at 88\degc{} also never throttles. THR misses 1.3 to 1.9
points fewer deadlines than it at 45\degc{} and 2.9 to 4.4 at 55\degc{}
(all seven traces), and has higher on-time utility on 6 and 7 traces,
while the thermostat keeps 0.8 to 1.6 points more accuracy. The
thermostat threshold is a dial, though. Sweeping it between 84 and 89\degc{}
(E8) traces an accuracy-timeliness curve, and THR lands on that curve,
between the 86 and 87\degc{} settings at 45\degc{} and at the 85\degc{}
setting at 55\degc. Fig.~\ref{fig:pareto} shows the same thing for duty
caps: every temperature-aware policy sits on the curve traced by fixed
C1 caps, and the matched-budget control, which spends THR's C1 share
with no temperature input, matches THR's miss rate on 6 of 7 traces at
55\degc{} and all 7 at 45\degc. At a single ambient and on the assumed
chip, the share of C1 frames determines the outcome, and the forecast
adds little beyond choosing that share well.

Choosing it well across temperatures is harder than it looks. A cap
tuned at 45\degc{} matches THR at 45\degc, then throttles 16\% of the
mission at 50\degc{} and 31\% at 55\degc, where THR misses 3.1 to 6.6
points fewer deadlines on every trace (Table~\ref{tab:caps}). Below
45\degc{} the same cap gives up 3.6 to 5.4 points of accuracy that THR
keeps. A cap tuned at 55\degc{} costs 7.1 points of accuracy at
45\degc{} and 12.1 at 35\degc. The matched budget also needed THR to
supply its number, and at 55\degc{} it still throttled up to 15.7\% of
the mission on three traces.

\begin{table}[t]
\caption{A C1 cap tuned once at 45\degc{} against THR at other ambients
(mean over traces). Thr.\ and Miss in \%.}
\label{tab:caps}
\centering\footnotesize
\setlength{\tabcolsep}{3.5pt}
\begin{tabular}{@{}lcccccc@{}}
\toprule
& \multicolumn{3}{c}{Cap tuned at 45\degc} & \multicolumn{3}{c}{THR} \\
\cmidrule(lr){2-4}\cmidrule(l){5-7}
Ambient & Thr. & Miss & Acc. & Thr. & Miss & Acc. \\
\midrule
35\degc & 0.0 & 39.5 & 0.946 & 0.0 & 49.9 & 1.000 \\
40\degc & 0.0 & 39.5 & 0.946 & 0.0 & 46.5 & 0.982 \\
45\degc & 0.0 & 39.5 & 0.946 & 0.0 & 39.6 & 0.950 \\
50\degc & 15.6 & 36.9 & 0.930 & 0.0 & 32.3 & 0.914 \\
55\degc & 31.0 & 31.2 & 0.902 & 0.0 & 25.1 & 0.883 \\
\bottomrule
\end{tabular}
\end{table}

\subsection{A different chip, with a calibrated forecast}
\label{sec:plant}
The thermostat's threshold and the caps were set for one plant. In E6
the true chip differs from Table~\ref{tab:params} and THR is given the
matching model, as if the chip had been characterized
(Table~\ref{tab:plant}). THR stayed at 0.0\% throttling on all five
variants, both temperatures and every trace. The 88\degc{} thermostat
throttled up to 29.0\% of the mission at 55\degc{} when $R_{\mathrm{th}}$
was 20\% higher, and up to 2.7\% when the chip heated twice as fast
($\tau\times0.5$). It reacts to the current reading, and a chip that
overshoots after the reading crosses 88\degc{} reaches 90\degc{} anyway.
THR looks 5\,s ahead and sees the overshoot coming. THR's miss
reduction over Belief was significant on all seven traces in every
variant except the cooler chip at 45\degc, where nothing throttled and
the difference was exactly zero. On the slow-heating chip
($\tau\times2$) at 45\degc{} Belief never throttled within the two
minutes, so THR's 3.7 to 4.4 point gain there comes from rationing C1
alone.

This comparison favours THR in one respect: THR gets the new chip's
$R_{\mathrm{th}}$ and $\tau$, and the thermostat keeps its 88\degc{}
threshold. The experiment measures what a calibrated forecast buys over
an uncalibrated threshold; whether THR copes with hardware it knows
nothing about remains untested. Calibration is cheap: one 7-minute heating curve
gave $R_{\mathrm{th}}$ and $\tau$ for the M5 (\S\ref{sec:m5}). The
thermostat could be retuned instead, but its right threshold depends on
both the chip and the ambient (\S\ref{sec:simple}), so it would need a
sweep for each pair. We did not run a retuned thermostat, and we did
not test an uncalibrated THR on the hotter chips at 55\degc. The
closest evidence is E4 (Table~\ref{tab:misspec}): at 45\degc, THR with
$R_{\mathrm{th}}$ off by 30\% in either direction never throttled,
and only its miss reduction changed.

\begin{table*}[t]
\caption{A different true chip (E6). THR uses the matching model.
$\Delta$ THR: reduction in miss rate relative to Belief (pp, range over
traces) and Holm-significant traces. Max thr.: largest throttled share
over traces (\%) for Belief, the 88\degc{} thermostat and THR.}
\label{tab:plant}
\centering\footnotesize
\setlength{\tabcolsep}{4pt}
\begin{tabular}{@{}l ccccc c ccccc@{}}
\toprule
& \multicolumn{5}{c}{Ambient 45\degc} && \multicolumn{5}{c}{Ambient 55\degc} \\
\cmidrule{2-6}\cmidrule{8-12}
True chip & $\Delta$ THR & Sig. & Belief & Tst. & THR && $\Delta$ THR & Sig. & Belief & Tst. & THR \\
\midrule
$R_{\mathrm{th}}\times0.8$ & +0.0 to +0.0 & 0/7 & 0.0 & 0.0 & 0.0 &  & +4.7 to +6.1 & 7/7 & 17.0 & 0.0 & 0.0 \\
$R_{\mathrm{th}}\times1.2$ & +6.7 to +7.9 & 7/7 & 28.4 & 0.0 & 0.0 &  & +8.2 to +12.1 & 7/7 & 51.8 & 29.0 & 0.0 \\
$\tau\times0.5$ & +7.9 to +9.6 & 7/7 & 15.8 & 0.0 & 0.0 &  & +11.0 to +14.7 & 7/7 & 42.3 & 2.7 & 0.0 \\
$\tau\times2$ & +3.7 to +4.4 & 7/7 & 0.0 & 0.0 & 0.0 &  & +4.8 to +5.6 & 7/7 & 25.5 & 0.0 & 0.0 \\
throttle power $0.8\times$ & +3.4 to +5.3 & 7/7 & 12.8 & 0.0 & 0.0 &  & +4.8 to +6.7 & 7/7 & 41.0 & 0.0 & 0.0 \\
\bottomrule
\end{tabular}
\end{table*}

\subsection{Physically consistent frame timing}
\label{sec:timing}
In E11 each decision lasts one deadline period $D$ of wall-clock time.
The thermal plant integrates over $\Delta t=D$ per decision. At each
decision the router takes the newest recorded frame, so decision $k$
uses the trace frame at time $kD$ and the frames in between are
skipped, as in a latest-frame robot pipeline. Inference still running
at the next boundary is abandoned, so work never overlaps, and THR's
horizon stays 5\,s of wall-clock time (15 to 24 decisions). Missions
are 1200 decisions, 4.2 to 6.8 minutes. Policies, seeds and the
88\degc{} thermostat are unchanged, and the 45\degc{} cap is retuned
under the new timing.

Table~\ref{tab:timing} shows the same picture, with larger effects.
Belief is throttled 14 to 17\% of the mission at 45\degc{} and 36 to
43\% at 55\degc, and THR is never throttled. Between 45 and 55\degc{}
THR misses 6.2 to 9.5 points fewer deadlines than Belief and 1.5 to 4.1
fewer than the thermostat, on every trace. Its on-time utility beats
Belief's on every trace too, at 15 to 18\% less energy per on-time
frame. Because each
decision now heats the chip by more, the 88\degc{} thermostat throttles
up to 4.9\% of the mission at 55\degc{} on the base chip, and up to
39\% on the chip with 20\% higher $R_{\mathrm{th}}$ and 17\% on the
chip with half the time constant, where THR stays at zero. The cap
tuned at 45\degc{} throttles 20 to 25\% of the mission at 50\degc. The
matched budget still ties THR on misses at every ambient, so the
budget explanation of \S\ref{sec:simple} carries over.

\begin{table}[t]
\caption{E11, consistent frame timing (one decision per $D$), mean
over traces. Thr.: \% of the mission throttled for Belief (B), the
88\degc{} thermostat (Tst), the cap tuned at 45\degc{} and THR.
$\Delta$: THR's miss reduction relative to Belief (pp, range over
traces). Last four columns: traces where THR is Holm-significantly
better / worse than each policy on miss rate (MB: matched budget).}
\label{tab:timing}
\centering\scriptsize
\setlength{\tabcolsep}{2.2pt}
\begin{tabular}{@{}lccccccccc@{}}
\toprule
& \multicolumn{4}{c}{Thr.\ (\%)} & & \multicolumn{4}{c}{THR vs} \\
\cmidrule(lr){2-5}\cmidrule(l){7-10}
Amb. & B & Tst & Cap & THR & $\Delta$ (pp) & B & Tst & MB & Cap \\
\midrule
35\degc & 0.0 & 0.0 & 0.0 & 0.0 & +0.0 to +0.1 & 0/0 & 0/7 & 0/0 & 0/7 \\
45\degc & 15.5 & 0.0 & 0.0 & 0.0 & +6.2 to +7.8 & 7/0 & 7/0 & 0/0 & 0/0 \\
50\degc & 27.6 & 0.0 & 23.0 & 0.0 & +8.2 to +9.5 & 7/0 & 7/0 & 0/0 & 7/0 \\
55\degc & 40.3 & 1.5 & 37.4 & 0.0 & +8.0 to +9.1 & 7/0 & 7/0 & 0/0 & 7/0 \\
\bottomrule
\end{tabular}
\end{table}

\subsection{When the deadline has slack}
\label{sec:slack}
With $D$ at C1's median, C1 misses about half its frames even on a cool
chip, so part of every policy's miss reduction comes from running C1
less. E5 removes that effect (Table~\ref{tab:deadline}). At
$1.25\times$ and $1.5\times$ the median, C1 meets the deadline on a cool
chip and only throttling causes misses. Belief then misses just 0.5\% of
frames at 45\degc{} and 1.0\% at 55\degc, even though it is throttled 12
and 36\% of the time, because C2 to C4 still finish in time on a slow
chip and Belief moves to them after one slow frame. THR's miss rate is
0.0\%, a small but significant difference on every trace. On-time
utility goes the other way: Belief is significantly better on 6 of 7
traces (0.970 against 0.950 at 45\degc), since THR's rationing of C1
costs more accuracy than throttling does. THR still uses 5\% less energy
per on-time frame (2.17 against 2.28\,J at 45\degc). With deadline
slack, THR is worth running only when throttling has costs outside
this metric, such as energy, a planner sharing the SoC, or wear.

\begin{table*}[t]
\caption{Deadline sweep (E5): means over 7 traces for Belief (B), the
88\degc{} thermostat (Tst) and THR. U is on-time utility. The last two
columns count traces where THR is Holm-significantly better / worse than
Belief on miss rate and on U.}
\label{tab:deadline}
\centering\footnotesize
\setlength{\tabcolsep}{4.5pt}
\begin{tabular}{@{}ll ccc ccc ccc cc@{}}
\toprule
& & \multicolumn{3}{c}{Miss (\%)} & \multicolumn{3}{c}{Throttled (\%)} & \multicolumn{3}{c}{U} & \multicolumn{2}{c}{THR vs B} \\
\cmidrule(lr){3-5}\cmidrule(lr){6-8}\cmidrule(lr){9-11}\cmidrule(l){12-13}
$D$ & Amb. & B & Tst & THR & B & Tst & THR & B & Tst & THR & miss & U \\
\midrule
1.0$\times$ & 35\degc & 49.9 & 49.9 & 49.9 & 0.0 & 0.0 & 0.0 & 0.501 & 0.501 & 0.501 & 0/0 & 0/0 \\
 & 45\degc & 44.3 & 41.2 & 39.6 & 12.2 & 0.0 & 0.0 & 0.532 & 0.546 & 0.554 & 7/0 & 5/0 \\
 & 55\degc & 32.3 & 29.0 & 25.1 & 36.3 & 0.0 & 0.0 & 0.607 & 0.610 & 0.632 & 7/0 & 6/0 \\
\midrule
1.25$\times$ & 35\degc & 0.0 & 0.0 & 0.0 & 0.0 & 0.0 & 0.0 & 1.000 & 1.000 & 1.000 & 0/0 & 0/0 \\
 & 45\degc & 0.5 & 0.0 & 0.0 & 12.2 & 0.0 & 0.0 & 0.970 & 0.958 & 0.950 & 7/0 & 1/6 \\
 & 55\degc & 1.0 & 0.0 & 0.0 & 36.3 & 0.0 & 0.0 & 0.919 & 0.899 & 0.883 & 7/0 & 1/6 \\
\midrule
1.5$\times$ & 35\degc & 0.0 & 0.0 & 0.0 & 0.0 & 0.0 & 0.0 & 1.000 & 1.000 & 1.000 & 0/0 & 0/0 \\
 & 45\degc & 0.5 & 0.0 & 0.0 & 12.2 & 0.0 & 0.0 & 0.970 & 0.958 & 0.950 & 7/0 & 1/6 \\
 & 55\degc & 1.0 & 0.0 & 0.0 & 36.3 & 0.0 & 0.0 & 0.919 & 0.899 & 0.883 & 7/0 & 1/6 \\
\bottomrule
\end{tabular}
\end{table*}

\subsection{Heat combined with camera faults}
\begin{figure}[t]
\centering
\includegraphics[width=\columnwidth]{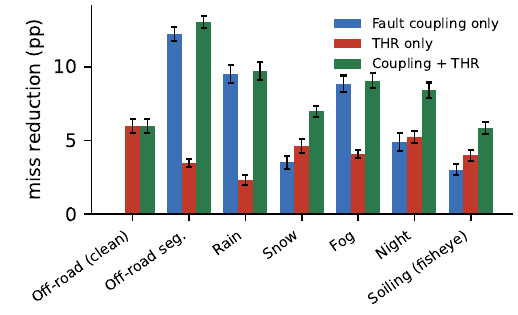}
\caption{Heat plus fault-coupled contention at 45\degc{} (E3): reduction
in miss rate relative to Belief, 95\% CIs over 20 seeds.}
\label{fig:compound}
\end{figure}

E3 adds fault-coupled contention on top of throttling
(Fig.~\ref{fig:compound}, Table~\ref{tab:compound}). The fault-coupling
term and THR respond to different causes. On the fault-free drive
Joint gains exactly 0.0 points, while THR gains 6.0 points at 45\degc{}
and 7.9 at 55\degc. At 25\degc{} THR adds nothing and Joint\,+\,THR
equals Joint exactly. At 45 and 55\degc{} Joint\,+\,THR beats Belief on
all seven traces, by 5.8 to 13.1 and 7.3 to 13.3 points, and is never
below Joint alone.

\begin{table}[t]
\caption{E3: reduction in miss rate relative to Belief (pp). J is
Joint; J+T is Joint\,+\,THR. $^\ast$: Holm-significant.}
\label{tab:compound}
\centering\footnotesize
\setlength{\tabcolsep}{3pt}
\begin{tabular}{@{}lcccccc@{}}
\toprule
& \multicolumn{2}{c}{25\degc} & \multicolumn{2}{c}{45\degc} & \multicolumn{2}{c}{55\degc} \\
\cmidrule(lr){2-3}\cmidrule(lr){4-5}\cmidrule(l){6-7}
Trace & J & J+T & J & J+T & J & J+T \\
\midrule
Off-road (clean) & +0.0 & +0.0 & +0.0 & +6.0$^\ast$ & +0.0 & +7.9$^\ast$ \\
Off-road seg. & +12.7$^\ast$ & +12.7$^\ast$ & +12.2$^\ast$ & +13.1$^\ast$ & +9.0$^\ast$ & +13.3$^\ast$ \\
Rain & +9.7$^\ast$ & +9.7$^\ast$ & +9.5$^\ast$ & +9.7$^\ast$ & +5.8$^\ast$ & +10.1$^\ast$ \\
Snow & +5.3$^\ast$ & +5.3$^\ast$ & +3.5$^\ast$ & +7.0$^\ast$ & +1.6$^\ast$ & +8.8$^\ast$ \\
Fog & +12.5$^\ast$ & +12.5$^\ast$ & +8.8$^\ast$ & +9.1$^\ast$ & +3.4$^\ast$ & +9.2$^\ast$ \\
Night & +8.0$^\ast$ & +8.0$^\ast$ & +4.9$^\ast$ & +8.4$^\ast$ & +2.4$^\ast$ & +9.6$^\ast$ \\
Soiling (fisheye) & +4.7$^\ast$ & +4.7$^\ast$ & +3.0$^\ast$ & +5.8$^\ast$ & +1.1$^\ast$ & +7.3$^\ast$ \\
\bottomrule
\end{tabular}
\end{table}

\subsection{THR's settings and a wrong model}
Because THR forecasts from the measured temperature, no setting we
tried brought throttling back in a meaningful way. Over $H\in\{20,50,100\}$
frames and $m\in\{0,2,4\}$\degc{} (E7), throttling stayed at 0.0\% except
for the most aggressive setting ($H=20$, $m=0$) at 55\degc, which
throttled up to 0.6\% of the mission. Longer horizons and larger margins
move THR down the same curve: at 45\degc{} the miss reduction over
Belief is between 0.5 and 11.0 points and mean agreement between 0.923
and 0.969, significant on all traces at every setting. A wrong thermal model
has the same effect (E4, Table~\ref{tab:misspec}). Underestimating
$R_{\mathrm{th}}$ by 30\% shrinks the miss gain to 0.8 to 1.8 points
and keeps more accuracy, and overestimating it does the opposite. THR
has one dial that matters, its effective conservatism, and getting the
model roughly right is enough to keep the chip out of throttling.

\begin{table}[t]
\caption{THR with a wrong thermal model at 45\degc{} (E4): range of the
miss reduction relative to Belief, Holm-significant traces, accuracy
range and the largest throttled share.}
\label{tab:misspec}
\centering\footnotesize
\setlength{\tabcolsep}{3.5pt}
\begin{tabular}{@{}lcccc@{}}
\toprule
THR's model & $\Delta$ miss (pp) & Sig. & Acc. & Max thr.\ (\%) \\
\midrule
true model & +3.9 to +5.4 & 7/7 & 0.917 to 0.994 & 0.0 \\
$R_{\mathrm{th}}\times0.7$ & +0.8 to +1.8 & 7/7 & 0.943 to 0.996 & 0.0 \\
$R_{\mathrm{th}}\times1.3$ & +7.1 to +9.0 & 7/7 & 0.888 to 0.993 & 0.0 \\
$\tau\times0.5$ & +5.6 to +7.6 & 7/7 & 0.899 to 0.994 & 0.0 \\
$\tau\times2$ & +3.2 to +4.4 & 7/7 & 0.925 to 0.995 & 0.0 \\
\bottomrule
\end{tabular}
\end{table}

\subsection{Bursts of misses under fault-coupled contention}
Re-scoring the original fault-coupled schedule of
\cite{roy2026beliefspace} (E1, Table~\ref{tab:bursts}), fault coupling
lowers the share of frames inside bursts of three or more misses
(0.3\,s) on 6 of 7 traces, by 0.6 to 5.6 points, and is exactly zero
on the clean drive. It shortens the single longest burst
significantly on only one trace.

\begin{table}[t]
\caption{Bursts under fault-coupled contention (E1). Burst share: \% of
frames in runs of at least 3 misses. Max run in frames.}
\label{tab:bursts}
\centering\footnotesize
\setlength{\tabcolsep}{3.5pt}
\begin{tabular}{@{}lccccc@{}}
\toprule
& \multicolumn{2}{c}{Burst share (\%)} & & \multicolumn{2}{c}{Max run} \\
\cmidrule(lr){2-3}\cmidrule(l){5-6}
Trace & Belief & Joint & $\Delta$ (pp) & Belief & Joint \\
\midrule
Off-road (clean) & 18.9 & 18.9 & +0.0 & 7.6 & 7.6 \\
Off-road seg. & 17.4 & 14.7 & +2.8$^\ast$ & 8.4 & 7.3 \\
Rain & 14.3 & 13.7 & +0.6$^\ast$ & 6.7 & 6.7 \\
Snow & 20.0 & 17.4 & +2.6$^\ast$ & 7.6 & 7.3 \\
Fog & 21.9 & 16.3 & +5.6$^\ast$ & 7.1 & 6.8 \\
Night & 19.7 & 16.9 & +2.8$^\ast$ & 7.8 & 7.2 \\
Soiling (fisheye) & 14.6 & 13.6 & +1.0$^\ast$ & 10.6 & 10.5 \\
\bottomrule
\end{tabular}
\end{table}

\subsection{Heating measured on a real chip}
\label{sec:m5}
\begin{figure}[t]
\centering
\includegraphics[width=\columnwidth]{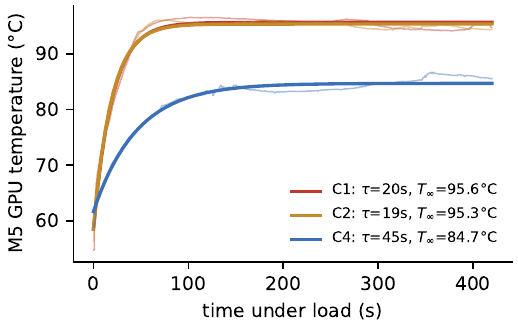}
\caption{Apple M5 GPU temperature under nonstop YOLO11 inference (thin)
with the fitted one-node RC step response (thick).}
\label{fig:m5}
\end{figure}

We had no Jetson, so we measured the Apple M5 laptop whose GPU produced
the profiled latencies (MacBook Pro, MPS, fp32, on mains power, other
apps closed apart from a stuck system process drawing about 7\,W of CPU
throughout). The \texttt{macmon} tool read GPU and CPU temperature,
power and fan speed four times a second without root access. For C1,
C2 and C4 we let the chip idle, ran the configuration nonstop for
7 minutes on $1024\times544$ frames, and fitted the step response of
Eq.~\eqref{eq:rc} to the GPU temperature (Table~\ref{tab:m5},
Fig.~\ref{fig:m5}).

A single RC node fits every curve with under 1\degc{} of error. For C1
the fitted time constant is 20\,s and the resistance 2.8\,\textdegree C/W,
close to the 30\,s and 2.0\,\textdegree C/W we assumed and inside the range
of chips tested in \S\ref{sec:plant}; nonstop C1 settles 41\degc{} above
idle, at 95.6\degc. The fit is a simplification. C4's curve has a
slower time constant (45\,s) and a higher resistance (3.3\,\textdegree C/W),
most likely because the fans speed up only under heavy load (5464\,rpm
after C1 against 2782 after C4), so the effective parameters depend on
the workload. The M5 never throttled: C1's latency was 187\,ms in the
first 30\,s and 186\,ms in the last 30\,s. This check therefore supports
the heating model and the scale of its parameters. It does not
reproduce throttling, which the fan-cooled laptop avoided.

\begin{table}[t]
\caption{Apple M5 under nonstop YOLO11 inference. Lat.: median latency
over the run, its first 30\,s and its last 30\,s (ms). $\Delta P$:
GPU+CPU power above idle (W). $\tau$ and $\Delta T$: fitted time
constant (s) and steady-state rise above idle (\degc). $R=\Delta T/\Delta P$
(\textdegree C/W). RMSE of the fit (\degc).}
\label{tab:m5}
\centering\footnotesize
\setlength{\tabcolsep}{2.6pt}
\begin{tabular}{@{}lcccccccc@{}}
\toprule
Cfg & Lat. & First & Last & $\Delta P$ & $\tau$ & $\Delta T$ & $R$ & RMSE \\
\midrule
C1 & 189 & 187 & 186 & 14.6 & 20 & 41.1 & 2.81 & 0.99 \\
C2 & 55 & 55 & 55 & 13.8 & 19 & 38.9 & 2.82 & 0.80 \\
C4 & 7 & 7 & 7 & 7.7 & 45 & 25.7 & 3.33 & 0.88 \\
\bottomrule
\end{tabular}
\end{table}

\section{Discussion and Limitations}

Reading the temperature is what keeps a router out of throttling;
routers that only watch latency throttle in every hot run we tried. At
one ambient on a known chip, the share of heavy-model frames sets the
accuracy-timeliness point, and a thermostat or duty cap at the right
setting does as well as THR. The forecast earns its place when the
ambient changes or the chip heats faster or settles hotter than the rule
assumed. In those runs the thermostat and the fixed caps throttled and
THR did not, which is the claim we make for it. On a chip that has
been characterized well, a tuned rule is just as good.

The deadline matters as much as the temperature. With the deadline at
the large model's median latency, avoiding throttling reduces misses
and energy. With 25\% slack, the cheaper models absorb throttling, and
a latency-only router delivers more on-time accuracy than THR.
Deployments should check which side of that line they are on.

The study is a simulation built on real camera traces and real
profiled latencies. The plant is one RC node with parameters we chose;
heating measured on an Apple M5 fits that model with parameters close
to ours, but the M5 never throttled, so throttling reuses the profiled
contended latency distribution instead of real DVFS states. The real throttle points reported for the
Jetson AGX Orin are higher (95.6 and 99.0\degc{} \cite{krislauks2026orin}),
and a lighter model in a ventilated enclosure did not throttle at
40\degc. Our plant variations cover part of that range, but only
hardware measurements can settle it. Latencies come from an Apple M5
GPU. Traces are repeated to reach mission length. Accuracy is
agreement with the largest model, which tracks real annotations poorly
\cite{roy2026beliefspace}. Seeds change noise on fixed frames and do not
stand in for independent missions.

Next we plan to measure $R_{\mathrm{th}}$, $\tau$ and throttle behaviour
on a Jetson Orin in an environmental chamber at 25 to 55\degc{} with
the included \texttt{tegrastats} harness, score the routers against
real annotations, and test whether a duty budget learned online from
temperature logs can replace the RC model.

\section{Conclusion}

On a hot robot, choosing the large detector now can cause throttling a
minute later. THR gives the router a closed-form temperature forecast
that costs microseconds per frame. With a tight deadline it kept the
chip out of throttling on all seven traces at 45 and 55\degc{} and cut
deadline misses by 3.9 to 7.5 percentage points and energy per on-time
frame by 12 to 13\% relative to latency-only routing. A tuned thermostat
reaches the same operating points on the chip it was tuned for, but
throttled for up to 29\% of the mission on a chip 20\% hotter than
assumed, where THR stayed at zero. With deadline slack, throttling is
cheap and THR's rationing is not worth its accuracy cost. For teams
building perception on hot hardware, the practical advice is to give
the router a temperature input, and to use a forecast when the chip's
thermal behaviour is uncertain.

\section*{Acknowledgment}
This work builds on the code and traces of belief-space perception
routing \cite{roy2026beliefspace}, led by Sparsh Roy as first author
with Vihan Aggarwal and Davin Yin. We thank Vihan Aggarwal and Davin
Yin for their part in that work.

\bibliographystyle{IEEEtran}
\bibliography{references}

\end{document}